\documentclass[lettersize,journal]{IEEEtran}
\usepackage{amsmath,amsfonts}
\usepackage{algorithmic}
\usepackage{algorithm}
\usepackage{array}
\usepackage{textcomp}
\usepackage{stfloats}
\usepackage{url}
\usepackage{verbatim}
\usepackage{graphicx}
\usepackage{cite}
\usepackage{enumerate}

\usepackage{multirow}
\usepackage{colortbl}
\usepackage{xcolor}

\begin{document}

\title{UAR-NVC: A Unified AutoRegressive Framework for Memory-Efficient Neural Video Compression}

% \thanks{This paper was produced by the IEEE Publication Technology Group. They are in Piscataway, NJ.

% This paper was produced by the IEEE Publication Technology Group. They are in Piscataway, NJ.

% This paper was produced by the IEEE Publication Technology Group. They are in Piscataway, NJ.}% <-this % stops a space

\author{Jia Wang, Xinfeng Zhang~\IEEEmembership{Senior Member,~IEEE}, Gai Zhang~\IEEEmembership{Student Member,~IEEE}, Jun Zhu~\IEEEmembership{Student Member,~IEEE}, Lv Tang~\IEEEmembership{Student Member,~IEEE} and Li Zhang~\IEEEmembership{Senior Member,~IEEE}
\thanks{This work was supported by the the Funds for International Cooperation and Exchange of the National Natural Science Foundation of China (Grant No. 62461160310), and the Key Program of the National Natural Science Foundation of China  (Grant No. 62431011). \textit{(Corresponding author: Xinfeng Zhang.)}

Jia Wang, Xinfeng Zhang, Gai Zhang, Jun Zhu and Lv Tang are with School of Computer Science and Technology, University of Chinese Academy of Sciences, Beijing, China (E-mail: \{wangjia242, xfzhang, zhanggai16, zhujun23\}@mails.ucas.ac.cn; luckybird1994@gmail.com). 

Li Zhang is with the Bytedance Inc., San Diego, USA (e-mail: lizhang.idm@bytedance.com).
}}

\markboth{Journal of \LaTeX\ Class Files,~Vol.~14, No.~8, August~2021}%
{Shell \MakeLowercase{\textit{et al.}}: A Sample Article Using IEEEtran.cls for IEEE Journals}

% \IEEEpubid{0}

\maketitle

\begin{abstract}
Implicit Neural Representations (INRs) have demonstrated significant potential in video compression by representing videos as neural networks. However, as the number of frames increases, the memory consumption for training and inference increases substantially, posing challenges in resource-constrained scenarios.
Inspired by the success of traditional video compression frameworks, which process video frame by frame and can efficiently compress long videos, we adopt this modeling strategy for INRs to decrease memory consumption, while aiming to unify the frameworks from the perspective of timeline-based autoregressive modeling.
In this work, we present a novel understanding of INR models from an autoregressive (AR) perspective and introduce a Unified AutoRegressive Framework for memory-efficient Neural Video Compression (UAR-NVC). UAR-NVC integrates timeline-based and INR-based neural video compression under a unified autoregressive paradigm. It partitions videos into several clips and processes each clip using a different INR model instance, leveraging the advantages of both compression frameworks while allowing seamless adaptation to either in form. 
To further reduce temporal redundancy between clips, we treat the corresponding model parameters as proxies for these clips, and design two modules to optimize the initialization, training, and compression of these model parameters. In special, the Residual Quantization and Entropy Constraint (RQEC) module dynamically balances the reconstruction quality of the current clip and the newly introduced bitrate cost using the previously optimized parameters as conditioning.
In addition, the Interpolation-based Initialization (II) module flexibly adjusts the degree of reference used during the initialization of neighboring video clips, based on their correlation.
UAR-NVC supports adjustable latencies by varying the clip length. Extensive experimental results demonstrate that UAR-NVC, with its flexible video clip setting, can adapt to resource-constrained environments and significantly improve performance compared to different baseline models. The project page: \url{https://wj-inf.github.io/UAR-NVC-page/}.

\end{abstract}


\section{INTRODUCTION}

\begin{figure}[t]
\centering
\includegraphics[width=0.5\textwidth]{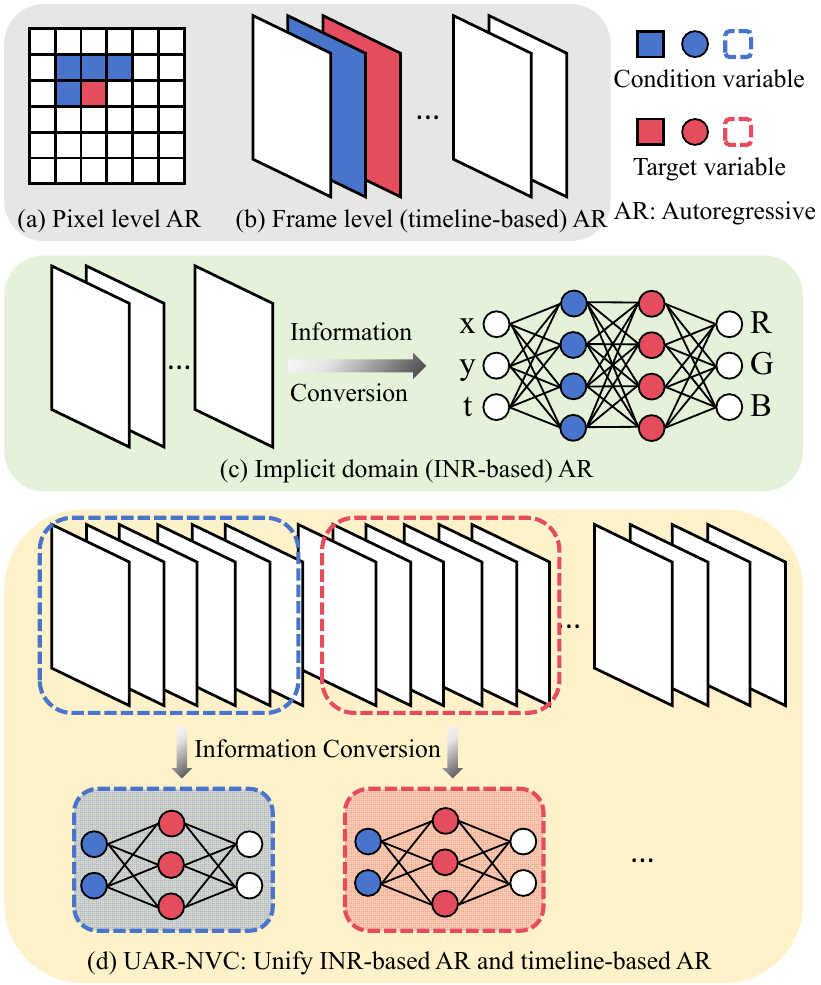}
\caption{Different AutoRegressive (AR) methods: (a) Pixel-level AR, which utilizes the PixelCNN~\cite{van2016pixel_pixelcnn} model to model pixel probabilities. (b) Frame-level AR, which employs techniques such as optical flow estimation to model relationships between frames. (c) Implicit domain (INR-based) AR, which first converts video information into a neural network and then performs AR between the layers of the feature map, considering coordinates, intermediate feature maps, and output values as variables, treating each network layer as a conditional probability model. (d) Our UAR-NVC, which integrates INR-based AR and timeline-based AR by using INR models to represent each video clip and applying AR between INR models to capture inter-clip relationships.}
\label{figure1}
\end{figure}

\IEEEPARstart{W}{ith} the advancement of media technologies, video data has become a widely used medium for connecting people to the world. However, this widespread use has also introduced significant challenges related to storage and transmission. To address these challenges, various video compression standards, such as H.264/AVC~\cite{H.264}, H.265/HEVC~\cite{H.265}, and H.266/VVC~\cite{H.266}, have been developed in recent decades. These standards generally treat video as an extension of images along the timeline and separately reduce the redundancies within and between frames.

Deep neural networks (DNNs) have recently achieved significant advancements across a wide range of fields. Consequently, neural video compression (NVC)~\cite{li2018fully, pfaff2018neural, hu2019progressive, lin2020m, chen2017deepcoder, wu2018video, toderici2017full, lu2019dvc, li2021dcvc} has evolved from replacing certain components~\cite{li2018fully, pfaff2018neural, hu2019progressive} of traditional compression frameworks to designing an end-to-end compression framework, such as DVC~\cite{lu2019dvc} in an autoencoder framework. This shift has significantly enhanced NVC’s potential, achieving superior rate-distortion performance through joint optimization.
Building on this, numerous studies have explored various advancements in individual modules, rapidly improving the Rate-Distortion (RD) performance of NVC. Although the most recent NVC codecs~\cite{li2023dcvc-dc} have surpassed VVC in RD performance, they still face challenges, including limited generalization capacity and suboptimal performance on unseen datasets~\cite{tang2025canerv}. Moreover, the autoencoder-style video compression frameworks encounter bottlenecks in decoding speed~\cite{chen2021nerv}, highlighting an urgent need for alternative approaches that simplify the pipeline and alleviate computational burdens.

In recent years, implicit neural representation (INR) has emerged as a powerful signal representation method, which has been used in image compression~\cite{dupont2021coin, dupont2022coin++, strumpler2022implicit, damodaran2023rqat, pham2023autoencoding} and video compression, such as NeRV~\cite{chen2021nerv}. NeRV utilizes the timestamp as its sole input, enabling the decoding of an entire frame in a single forward pass, thereby significantly accelerating the decoding speed. Numerous follow-up works based on NeRV~\cite{li2022e-nerv, lee2023ffnerv, chen2023hnerv, kwan2024hinerv, zhang2024boosting, maiya2023nirvana, he2023d-nerv, tang2023scene} have further improved the performance of INR codecs. Among these, HiNeRV~\cite{kwan2024hinerv} demonstrated notable improvements in RD performance for videos with obvious dynamic content, while MVC~\cite{tang2023scene} even outperformed H.266/VVC by 20\% in BD-Rate~\cite{bjontegaard2001BDBR} saving when tested on conference and surveillance video datasets. 
Despite these advancements, INR-based approaches face significant challenges due to the reliance on a single model to encode all video frames, which places a heavy burden on GPU memory during both training and inference. This issue renders them impractical for many applications. How to decrease the training burden or processing longer video becomes a significant problem. An intuitive solution is to divide the video into clips, and use multiple smaller INR networks. However, this strategy will also introduce redundancy among models for different clips, potentially degrading the rate-distortion performance of INR-based video compression systems.

To address this issue, we review the representative traditional framework H.264/AVC~\cite{H.264}, which compresses videos frame by frame using concepts like I-frames and P-frames. This enables efficient processing of long videos with limited memory resources. This raises a key question: can a similar concept be applied when using INRs to model videos? The traditional I-frame and P-frame modeling approach can be interpreted as an autoregressive process over time, where information from previous frames helps model subsequent ones. In this process, each frame serves as a fundamental unit. More specifically, video coding consists of intra-frame and inter-frame encoding, with the latter conceptualized as a form of timeline autoregression. Extending this idea, we propose expanding timeline autoregression from the frame level to the clip level.

Building on this concept, we divide a video into multiple clips, where each clip is modeled using a dedicated instance of an INR model, trained in an autoregressive manner. The clip length can be configured by users based on their available computational resources. 
To leverage the correlation between clips, we propose the Interpolation-based Initialization (II) module, which improves model initialization efficiency by utilizing information from INR models trained on previous clips.
To avoid introducing redundant information between neighboring INR models, we design the Residual Quantization and Entropy Constraint (RQEC) module, which dynamically balances restoration quality and newly introduced information. In addition, to support random access, we introduce the concept of the Group of Models (GOM), similar to the Group of Pictures (GOP) in traditional video coding. However, unlike GOP, GOM treats clips, rather than frames, as the fundamental units in timeline autoregression. 
Building on these modules, we propose a novel Neural Video Compression (NVC) framework that combines the powerful modeling capabilities of INRs with the flexibility of traditional video coding frameworks to exploit the correlation of longer time domains, termed the Unified AutoRegressive-based Neural Video Compression framework (UAR-NVC).

The contributions of our work are summarized as follows:
\begin{itemize}
    
    \item Unified Framework: We propose UAR-NVC, a novel framework that unifies timeline-based and INR-based NVC frameworks. Most existing INR model can be seamlessly integrated into our framework, demonstrating its flexibility and generality.
    
    \item Enhancements in Autoregressive Modeling: We design two modules to enhance autoregressive modeling for videos under the proposed UAR-NVC framework. The RQEC module dynamically balances the newly introduced information and the distortion of the current clip, based on the INR model parameters of the previous clip. The II module adaptively controls the reference strength for INR model initialization, allowing it to handle diverse video data effectively.

    \item Comprehensive Experiments: We conduct extensive experiments to verify the efficiency of the proposed UAR-NVC framework. The results demonstrate its superior performance in rate-distortion optimization and highlight its advantages for practical applications.

\end{itemize}

\begin{figure*}[ht]
\centering
\includegraphics[width=1.0\textwidth]{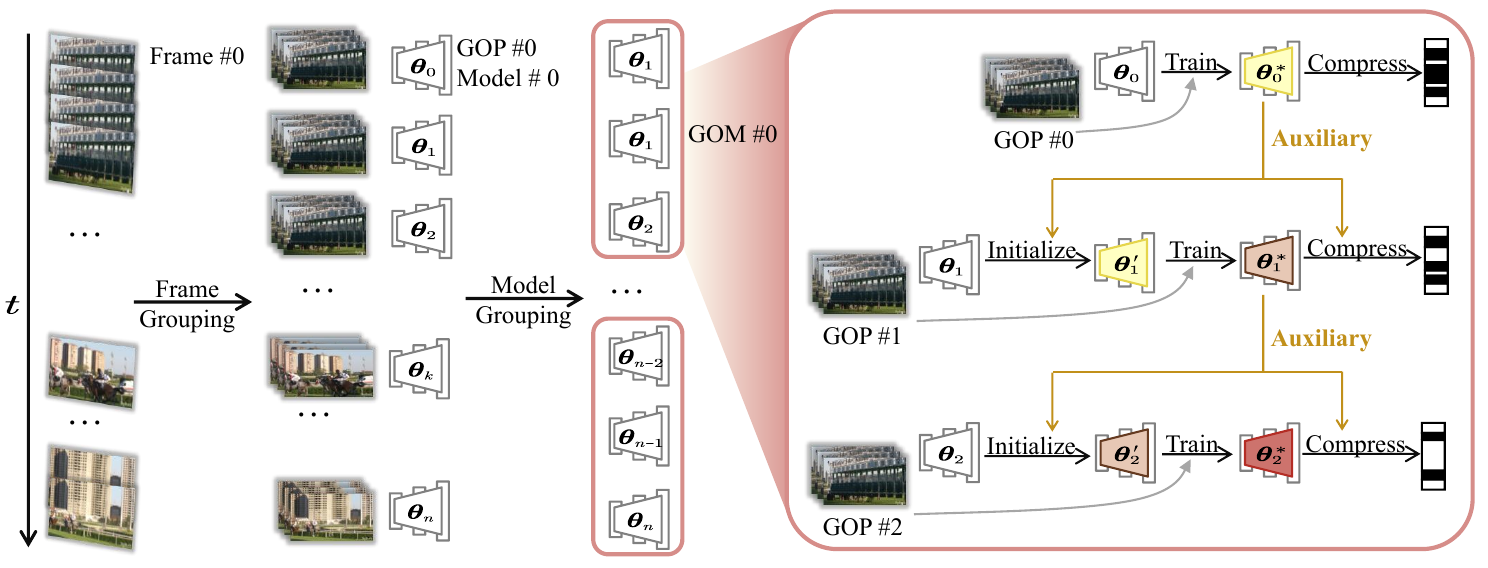}
\caption{The proposed UAR-NVC, a practical INR framework for video compression. The video frames are grouped into several GOPs (video clips), where we train one model for one GOP. To balance correlation capture and random access, GOPs are grouped into GOMs. And time dependency exists only between GOPs within a GOM. The right part shows the training and compression pipeline of one GOM in UAR-NVC.}
\label{fig-UAR-NVC}
\end{figure*}

\section{RELATED WORK}

\subsection{Traditional Video Compression Standards}

As early as 1969, video coding was proposed as an academic problem at the Image Coding Workshop in Boston. Subsequently, under the impetus of the Telecommunication Standardization Organization of the International Telecommunication Union (ITU~\cite{ITU}), it rapidly developed into the H.261 standard~\cite{H.261}.
As the earliest video coding standard, H.261 prescribed all components of video coding, including motion compensation, DCT transform, quantization, and entropy coding. Following this, H.262~\cite{H.262}, H.263~\cite{H.263}, and H.264/AVC~\cite{H.264} standards were proposed successively. With the introduction of multi-frame reference, intra-frame prediction, and multi-scale coding block technology, H.264 became the most widely adopted coding standard, marking the maturity of the hybrid coding framework. To further improve performance, H.265/HEVC~\cite{H.265} was proposed, achieving the same visual quality with nearly half the bitrate compared to H.264. More recently, the H.266/VVC~\cite{H.266} standard has emerged, offering even better performance.

Besides the H.26x series, the AVS series developed by the Audio and Video Coding Standard Workgroup of China~\cite{AVS2023} has also gained significant attention. Meanwhile, the video coding standards VP8~\cite{VP8}, VP9~\cite{VP9}, and AV1~\cite{AV1}, led by Google, continue to evolve. Despite these advancements, nearly all of these standards follow the pipeline established by H.264, which underscores the robustness of the hybrid video coding framework. This highlights the stability and practicality of this framework, which served as an inspiration for our work.

\subsection{Neural Video Compression}
With the popularity of deep learning, many works~\cite{zhang2017effective, li2018fully, pfaff2018neural, hu2019progressive} have focused on designing neural modules to replace certain components in traditional video coding frameworks. \cite{li2018fully} proposes a fully connected network for intra prediction. \cite{pfaff2018neural} proposes training multiple networks as different prediction modes. \cite{hu2019progressive} further incorporates RNNs and CNNs into the design of intra prediction networks, respectively. \cite{yan2017convolutional} and \cite{yan2018convolutional} optimize motion compensation using an improved subpixel interpolation algorithm. Meanwhile, \cite{lin2020m} introduces a multi-frame reference and bidirectional motion estimation mechanism, effectively improving the accuracy and compression efficiency of inter-frame prediction. In the aspect of entropy models, researchers have used CNNs or MLPs to predict the probability distribution of different data parts~\cite{lin2020m, song2017neural, ma2018convolutional}.

In another aspect, several researchers have proposed new video coding schemes~\cite{chen2017deepcoder, wu2018video, lu2019dvc, wang2022end, chen2023sparse, li2024neural, liu2024enlarged}. For instance, \cite{chen2017deepcoder} trains two autoencoders using deep neural networks to compress intra-frame blocks and the inter prediction residuals for I-frames or P-frames, respectively. \cite{wu2018video} employs the image coding scheme introduced in \cite{toderici2017full} to compress I-frames, while utilizing image interpolation to compress motion information from known frames for the remaining frames. 
\cite{lu2019dvc} proposes the first end-to-end neural video compression framework by replacing all modules in the traditional framework and enabling joint optimization across the entire system. Leveraging the powerful learning ability of neural networks, DCVC~\cite{li2021dcvc} uses a conditional method to capture the correlation between the current frame and the reference frame instead of simply using residuals. The subsequent work, 
% DCVC-DC~\cite{li2023dcvc-dc}, further enhances neural video compression (NVC) by incorporating diverse contexts. 
DCVC-FM~\cite{li2024neural}, further enhances Neural Video Compression (NVC) by introducing feature modulation techniques and support a wide quality range, and can effectively handle long prediction chains.
Despite these advancements, these methods face generalization challenges and cannot get the optimal latent, as their model parameters are amortized over a fixed training dataset. To address this limitation, a new learning approach that can optimize model parameters specifically for each individual video is needed.

\subsection{INR for Image Compression}

Recently, Implicit Neural Representation (INR) has gained significant attention and has been successfully applied in the representation of objects~\cite{park2019deepsdf}, voxel~\cite{song2024spc} and scenes~\cite{sitzmann2019scene}. COIN~\cite{dupont2021coin} introduces INR into image compression. In this framework, the target image is represented as a set of pixels, where each pixel is defined by its coordinates and color values: $(x, y, R, G, B)$. 
By learning the mapping function $f_{\boldsymbol{\theta}}:\left( x,y \right) \rightarrow \left( R,G,B \right)$ parameterized by the network parameters $\boldsymbol{\theta}$, the information of the image is encoded into the network parameters. Afterward, COIN applies quantization and entropy coding to these parameters and stores them in a binary file, completing the encoding process. The decoding process involves multiple forward passes of the network for all the pixels, following entropy decoding and dequantization of the parameters.

Subsequently, several works have focused on reducing the encoding time through meta-learning~\cite{dupont2022coin++} and improved rate-distortion (RD) performance via better network architectures~\cite{catania2023nif} or more precise quantization techniques~\cite{damodaran2023rqat, gordon2023quantizing}. Other works have further enhanced the RD performance through model pruning\cite{schwarz2023modality} or by combining autoencoders~\cite{pham2023autoencoding}. COOL-CHIC~\cite{ladune2023cool} innovatively combines INR with feature embedding, proposing a low-complexity image codec. 
% Building on COOL-CHIC, the subsequent work C3~\cite{kim2023c3} demonstrated RD performance close to VTM on the UVG video benchmark with less than 5k MACs/pixel for decoding.
Building on COOL-CHIC, the subsequent work C3~\cite{kim2023c3} further optimized its quantization strategies and model architecture, demonstrating RD performance close to VTM on the UVG video benchmark with less than 5k MACs/pixel for decoding. Besides, C3 models the video as 3D patches to mitigate the large memory footprint associated with full-video processing.
This highlights the promising advantages of implicit neural representation methods, particularly in resource-constrained environments.

\subsection{INR for Video Compression}

Different from coordinate-based INR models like SIREN~\cite{sitzmann2020implicit} and COIN, NeRV~\cite{chen2021nerv} uses the time stamp $t$ as the input of the INR network, allowing it to reconstruct a entire video frame in one forward pass. This approach can achieve faster decoding speeds even than the practical traditional codec H.264. 
IPF~\cite{zhang2021implicit} using the explicit relation of two frames (I-frame and P-frame) to facilitate modeling.
E-NeRV~\cite{li2022e-nerv} combines spatial coordinates with time stamps and replaces the convolution kernel with two consecutive convolution kernels with smaller channels. 
PS-NeRV~\cite{bai2023ps-nerv} reconstructs patches of the image instead of the whole image and uses Adaptive Instance Normalization (AdaIN) to modulate the features of the later convolution layers according to a time-coordinate condition. 
FF-NeRV~\cite{lee2023ffnerv} uses learnable multi-scale resolution feature grids to replace time or spatial coordinates, incorporates flow information into frame-wise representations to exploit the temporal redundancy across frames, and introduces quantization-aware training to shrink the quantization gap between training and model compression. 
HNeRV~\cite{chen2023hnerv} trains an encoder to generate content-adaptive embeddings and proposes evenly distributed parameters for HNeRV, providing more capacity to store high-resolution content and video details.
DSNeRV~\cite{yan2024ds-nerv} decomposes video into static part and dynamic part, using two group embedding to represent each.
T-NeRV~\cite{saethre2024combining} decomposes embedding into frame level and GOP level, and use the encoder in HNeRV to learning better frame embedding.
HiNeRV~\cite{kwan2024hinerv} uses the learnable feature grid not only at the network input but also in the subsequent upsampling module. By further improving the pruning and quantization parts of the model compression pipeline, HiNeRV achieves a $43.4\%$ overall bit rate saving over DCVC on the UVG dataset, measured in PSNR. 
Boost-NeRV~\cite{zhang2024boosting} proposes three universal techniques for NeRV-style models in module conditioning, activation layers, and entropy constraints, respectively.

Besides these excellent works that progressively improve the rate-distortion performance of Video INR, some works focus on the application of Video INR. To handle videos with frequent scene switching, D-NeRV~\cite{he2023d-nerv} represents a large and diverse set of videos as a single neural network by using a multi-tiered structure together with a Visual Content Encoder and Motion-aware Decoder. Additionally, MVC~\cite{tang2023scene} employs several spatial context enhancement and temporal correlation capturing techniques to further improve the representation capability of Video INR, achieving up to a $20\%$ bitrate reduction compared to the latest video coding standard H.266 in conference and surveillance videos. PNVC~\cite{gao2025pnvc} introduces a pre-trained image encoder to accelerate the encoding process. To process long and different resolution videos, NIRVANA~\cite{maiya2023nirvana} proposes fitting every continuous three frames using different independent patch-wise model instances and employs an autoregressive approach in model initialization and compression. 
However, NIRVANA is limited by its specific model structure, we unify the timeline-based framework and INR-based NVC framework from an autoregressive perspective. Our UAR-NVC framework allows users to choose different GOP sizes based on available resources, supports various INR base models, and includes two modules that significantly leverage the correlation between clips, thereby greatly improving rate-distortion (RD) performance for video compression.

\section{PROPOSED METHOD}
\label{sec:proposed method}

\begin{algorithm}[thb]
    \caption{Pipeline of UAR-NVC}
    \label{algorithm0}
\begin{algorithmic}

    \STATE {\bfseries procedure} Video Process($\mathbf{V},T,p,m$)
    % \STATE GOP\_nums $\gets$ T / p
    \STATE $GOM\_nums \gets T / (p \cdot m)$
    \FOR{$i_{gom}=0$ {\bfseries to} $GOM\_nums-1$}
        \STATE $\left( i_s,i_e \right) =\left( i_{gom},i_{gom}+1 \right) \cdot p\cdot m$
        \STATE $GOPs:=\mathbf{V}\left[ i_s:i_e \right]$
        \STATE $INR\_TASK\left( GOM,p,m \right)$
    \ENDFOR

\STATE

    \STATE \textit{// Training process within a GOM}
    \STATE {\bfseries procedure} $INR\_TASK\left( GOM,p,m \right)$ 

    \FOR{$k=0$ {\bfseries to} $m-1$}
        \IF{$k==0$}
            \STATE $\boldsymbol{\theta}_{k} \gets Normal(\boldsymbol{0}, \boldsymbol{I})$
            \STATE $\boldsymbol{\theta}_{k}^{\prime} \gets \boldsymbol{\theta}_{k} $
        \ELSE
            \STATE $\boldsymbol{\theta}_{k}^{\prime} \gets f_i \left( \boldsymbol{\theta}_{k-1}, \boldsymbol{\theta}_{k-1}^* \right)$
        \ENDIF

        \STATE $GOP := GOPs\left[ k\cdot p:k \cdot (p + 1) \right]$
        
        \FOR{$e=0$ {\bfseries to} $epochs$}
            \FOR{$i_{frame}=0$ {\bfseries to} $p-1$}
                \STATE $\mathrm{v}_{frame} := GOP\left[ i_{frame}\right]$
                \STATE $\hat{\mathrm{v}}_{frame} \gets \mathrm{NeRV}.forward\left( t;\boldsymbol{\theta }_{k}^{*} \right) $ 
                \STATE $loss_d \gets MSE\left( \mathrm{v}_{frame},\hat{\mathrm{v}}_{frame} \right) $
                \STATE $loss_r \gets I\left( f_c\left( \boldsymbol{\theta }_{k}^{\prime},\boldsymbol{\theta }_{k}^{*} \right) \right) $
                \STATE $\left( loss_r+ \lambda \cdot loss_d \right) .backward\left(  \right) $ 
            \ENDFOR
        \ENDFOR
        \STATE $bitstream \gets f_c\left( \boldsymbol{\theta }_{k}^{\prime},\boldsymbol{\theta }_{k}^{*} \right)$
                
    \ENDFOR

% \EndProcedure
\end{algorithmic}
% \# The functions `forward()" and `backward()" represents the forward process of network and backward process of loss.
\end{algorithm}

To enhance the practicality of INR models for video compression tasks while ensuring compatibility with existing models, we propose the UAR-NVC framework from an autoregressive perspective, as illustrated in Fig.~\ref{fig-UAR-NVC}. In Sec.~\ref{sec:Information Autoregressive}, we review the advancements in autoregressive (AR) modeling for timeline-based video compression and provide a novel interpretation of INR models. In Sec.~\ref{The Proposed UAR-NVC Framework}, we introduce our UAR-NVC framework, followed by a detailed explanation of its three key components. Sec.~\ref{sec-Group of Models} discusses the detailed methodology for segmenting an input video. In Sec.~\ref{sec-Residual QAT and Entropy Constrain}, we present the Residual Quantization and Entropy Constrain (RQEC) module. Finally, in Sec.~\ref{sec-Interpolation-based Model Initialization}, we describe the model initialization process within UAR-NVC.

\subsection{Information Autoregressive}
\label{sec:Information Autoregressive}

In the context where $x$ represents a pixel, PixelCNN~\cite{van2016pixel_pixelcnn} can be employed to model spatial relationships through the probability distribution $p(x_i | x_{i-1}x_{i-2}...)$, which utilizes preceding pixels to model the probability distribution of the current pixel. This approach is referred to as pixel-level AR. If $x$ denotes a frame, temporal relationships can be modeled using $p(x_i | x_{i-1})$ by leveraging motion estimation between neighboring frames~\cite{li2021dcvc}. This method captures temporal correlations within a video and is referred to as frame-level AR. Unlike generation tasks, compression tasks require recording both the signal distribution and the sampling position, as accurate signal recovery is necessary. 
Therefore, we consider the AR process in compression as an alternating process: initially obtaining the probability distribution, subsequently introducing new information decoded from the bitstream, and finally determining the value of the intermediate variable, which gradually converges to the original signal.

For an INR model, training is conducted over multiple epochs to encode information from images or videos into the network parameters. The reconstructed image or video is then obtained through a forward pass. Considering the network's input (coordinates), the output image or video, and the intermediate features as variables, the forward process can be interpreted as estimating a probability distribution. This process introduces new information from the bitstream into the network parameters, computes the next layer's feature values, and progressively refines the reconstructed frame, akin to the autoregressive (AR) modeling approach previously discussed.
The two primary differences between timeline-based and INR-based compression models are as follows: (1) Timeline-based models store information transformation exclusively in the latent space, whereas INR-based models encode it within the network parameters. (2) During video decoding, conventional timeline-based models introduce information from the bitstream and perform the transformation in a single pass, whereas INR-based models repeat this process multiple times.

Since the transformation within a single convolutional or multilayer perceptron (MLP) layer is relatively simple, deeper network architectures are often necessary for mapping coordinates to images or videos. However, increasing network depth substantially raises GPU memory consumption. To address this issue, we partition the video into clips and model each clip using a separate INR model instance, thereby reducing memory usage. In addition, we apply timeline-based autoregressive (AR) modeling by treating each clip as a symbol and modeling them sequentially to further minimize temporal redundancy. We refer to this approach as Unified AR.

\subsection{The Proposed UAR-NVC Framework}
\label{The Proposed UAR-NVC Framework}

We begin by introducing the foundational concepts of our framework. A video with $T$ frames can be represented as $\mathbf{V}=\left\{ \mathrm{v}_t \right\} _{t=1}^{T}\in \mathbb{R} ^{T\times H\times W\times 3}$. The INR model can be regarded as a mapping function $f_{\boldsymbol{\theta}}:\mathbb{R} \rightarrow \mathbb{R} ^{H\times W\times 3}$ parameterized by $\boldsymbol{\theta}$. Given a time stamp $t$, the reconstructed frame $\hat{\mathrm{v}}_t$ is obtained, forming the complete reconstructed video $\hat{\mathbf{V}}=\left\{ \hat{\mathrm{v}}_t \right\} _{t=1}^{T}\in \mathbb{R} ^{T\times H\times W\times 3}$. 
We define $p$ and $m$ as user-defined hyperparameters representing the size of GOP (Group of Pictures) and GOM (Group of Models), respectively. The video is partitioned based on these parameters. Each GOP instance is denoted as GOP$\#k, k \in \mathbb{R}$, and each GOM instance as GOM$\#k, k \in \mathbb{R}$, where $k$ represents the index of the respective GOP or GOM. The first model in a GOM is referred to as the I-model, meaning it does not depend on any preceding model. The remaining models in a GOM are termed P-models, as they reference the previous model for initialization and compression.

As illustrated in Fig.~\ref{fig-UAR-NVC}, we begin by grouping video frames into GOPs based on the parameter $p$. Each GOP contains the same number of frames and follows an identical model structure, but distinct model instances are assigned to each GOP. We employ an INR model, such as HNeRV~\cite{chen2023hnerv}, to fit the video frames within each GOP. Subsequently, GOPs are further grouped into GOMs according to the parameter $m$. All GOMs follow a consistent processing pipeline, where INR models are sequentially trained using their corresponding GOPs as training data.

Within a GOM, we employ an autoregressive approach for training and compressing each model. For Model$\#0$, initialized with randomly parameters $\boldsymbol{\theta}_{0}$, the frames in GOP$\#0$ serve as the training data. After training, the parameters are optimized from $\boldsymbol{\theta}_{0}$ to $\boldsymbol{\theta}_{0}^{*}$, denoted as: $\boldsymbol{\theta }_0\xrightarrow{train}\boldsymbol{\theta }_{0}^{*}$. The corresponding bitrate is defined as:
\begin{equation}
    Rate_0=I\left( \boldsymbol{\theta }_{0}^{*} \right), 
\end{equation}
where $I(\cdot)$ calculates the total bitrate required to encode the given parameters.

Since the randomly initialized parameters $\boldsymbol{\theta}_{0}$ can be reconstructed using the initialization seed, and $\boldsymbol{\theta}_{0}^{*}$ can be read back from the bit stream, both (primarily $\boldsymbol{\theta}_{0}^{*}$) serve as reference information for processing Model$\#1$. The relationship between adjacent GOPs is captured through model parameters. Specifically, instead of using randomly initialized parameters $\boldsymbol{\theta}_{1}$, we derive specially initialized parameters  $\boldsymbol{\theta}_{1}^{\prime}$ for Model$\#1$ by combining $\boldsymbol{\theta}_{0}$ and $\boldsymbol{\theta}_{0}^{*}$, defined as:
\begin{equation}
    \boldsymbol{\theta }_{1}^{\prime}=f_i\left( \boldsymbol{\theta }_0,\boldsymbol{\theta }_{0}^{*} \right).
\end{equation}
This initialization strategy accelerates the training of Model$\#1$, with experimental analyses provided in Sec.~\ref{sec:experiments}. Furthermore, since the final parameters $\boldsymbol{\theta}_{1}^{*}$ are obtained through training (or fine-tuning) from the primary parameters $\boldsymbol{\theta}_{1}^{\prime}$, denoted as: $\boldsymbol{\theta }_{1}^{\prime}\xrightarrow{train}\boldsymbol{\theta }_{1}^{*}$, we store the difference between $\boldsymbol{\theta}_{1}^{\prime}$ and $\boldsymbol{\theta}_{1}^{*}$ instead of directly recording $\boldsymbol{\theta}_{1}^{*}$. The bitrate of Model$\#1$ is given by:
\begin{equation}
    Rate_1=I\left( f_c\left( \boldsymbol{\theta }_{1}^{\prime},\boldsymbol{\theta }_{1}^{*} \right) \right),  
\end{equation}
where the definition of $f_c()$ will be given in Sec.~\ref{sec-Residual QAT and Entropy Constrain}. The remaining P-models follow the same reference process, differing only in the reference target. The complete processing pipelines is detailed in Alg.~\ref{algorithm0}.

\subsection{Group of Models}
\label{sec-Group of Models}

\begin{figure}[t]
\includegraphics[width=0.5\textwidth]{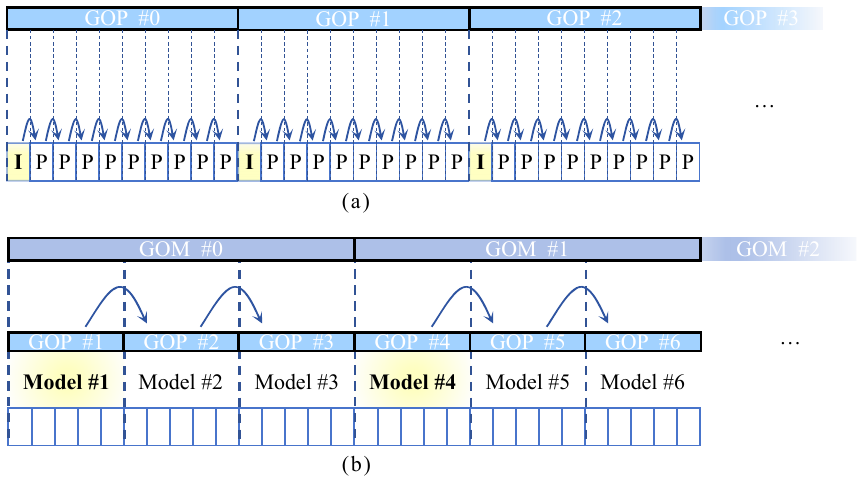}
\caption{(a): GOP level video partition with I-frames and P-frames. (b): GOM level video partition with I-models(\#1/4) and P-models(\#2/3/5/6).}
\label{fig-GOM}
\end{figure}

Traditional video compression standards, such as H.264, categorize video frames into three types: intra-coded (I) frames, predictive-coded (P) frames, and bidirectionally predicted (B) frames, to regulate the reference direction of each frame. The Group of Pictures (GOP) structure segments a video sequence into multiple groups, where frames within a GOP can reference each other while ensuring a one-way reference direction between any two frames. Across different GOPs, compression and decompression occur independently, enabling resynchronization during decoding in practical applications such as video jump playback.

Inspired by this setting, we define the set of frames which are modeled together a GOP, equivalent to a clip in our framework. Consequently, each GOP is assigned a separate INR model instance, as described in Sec.~\ref{The Proposed UAR-NVC Framework}.
Building on this concept, we extend the traditional GOP definition from the frame level to the model level, introducing the Group of Models (GOM). Analogous to GOPs, we designate the first model in a GOM as an I-model, while subsequent models are classified as P-models. The I-model is initialized and compressed independently, whereas P-models rely on the parameters of the preceding model for initialization and compression.
Under this framework, a given video sequence is first divided into multiple GOPs through frame grouping, followed by further partitioning into GOMs through model grouping. We use equal-length and continuous sub-items (frames/models) in this process, though it can be extended to uneven or dynamic partitioning based on sequence variability or user-defined requirements. The partitioning and reference process is illustrated in Fig.~\ref{fig-GOM}.

\begin{figure}[t]
\includegraphics[width=0.5\textwidth]{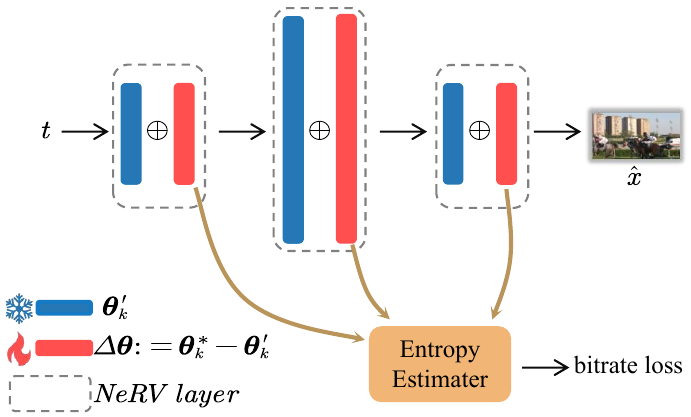}
\caption{Residual quantization-based entropy estimation on P-model. $\boldsymbol{\theta } _{k}^{\prime}$ denotes the parameters initialized from the model trained previously and $\boldsymbol{\theta }_{k}^{*}$ denotes the parameters used for the forward process of network. Since the model from the same GOM is accessible at the initialization of the current model, we can dynamically capture the relationship between these two models by estimating the entropy of $\varDelta \boldsymbol{\theta}$ during training. Notice that $\varDelta \boldsymbol{\theta}$ is the residual of $\boldsymbol{\theta }_{k}^{*}$ and $\boldsymbol{\theta } _{k}^{\prime}$ in our deployment, though alternative formats can also be employed.}
\label{fig-RQAT}
\end{figure}

\subsection{Residual Quantization and Entropy Constrain}
\label{sec-Residual QAT and Entropy Constrain}

COIN~\cite{dupont2021coin} and NeRV~\cite{chen2021nerv} reformulate video compression as a model compression problem by fitting videos using neural networks. Many subsequent works adopt the model compression pipeline introduced by NeRV, which consists of model pruning, model quantization, and parameter encoding.
Among these, the latest INR model, HiNeRV~\cite{kwan2024hinerv}, introduces several advancements. It applies a higher pruning ratio to convolutional layers with wider channels, based on the assumption that they contain more redundancy. In addition, it leverages Quantization-Aware Training (QAT) to mitigate the quantization gap between training and inference. Furthermore, HiNeRV replaces the Huffman encoder with a more advanced arithmetic encoder, significantly improving rate-distortion performance.

In our view, certain works~\cite{gomes2023video, maiya2023nirvana, zhang2024boosting} can further improve compression performance by dynamically balancing reconstruction quality and bitrate cost. This can be achieved by estimating the entropy of network parameters and incorporating it into the training loss. Therefore, we adopt the entropy-constrained method from Boost-NeRV~\cite{zhang2024boosting}, in which a symmetric scalar quantization scheme with a trainable scale parameter $\varsigma$ is used for model parameters:
\begin{equation}
    Q(\boldsymbol{\theta})=\left\lfloor\frac{\boldsymbol{\theta}}{\varsigma}\right\rceil, \quad P(\boldsymbol{\theta})=\boldsymbol{\theta} \times \varsigma, 
\end{equation}
where $Q(\cdot)$ represents scale and quant, while $P(\cdot)$ represents scale back. Furthermore, we model the probability distribution of the quantized parameters $\hat{\boldsymbol{\theta}}$ using a Gaussian distribution:
\begin{equation}
p\left(\hat{\boldsymbol{\boldsymbol{\theta}}}\right)=\prod_{i}\left(\mathcal{N}\left(\mu_{\boldsymbol{\boldsymbol{\theta}}}, \sigma_{\boldsymbol{\boldsymbol{\theta}}}^{2}\right) * \mathcal{U}\left(-\frac{1}{2}, \frac{1}{2}\right)\right)\left(\hat{\boldsymbol{\theta}}_{i}\right),
\end{equation}
where $\mu_{\boldsymbol{\boldsymbol{\theta}}}$ and $\sigma_{\boldsymbol{\boldsymbol{\theta}}}^{2}$ are the mean and variance of the parameters $\boldsymbol{\theta}$ for each layer, and $i$ represents layer index. Here, $*$ denotes convolution. As previously mentioned, this approach captures global relationships among all parameter elements. Notably, we observed that parameters in wider convolutional layers are allocated fewer bits, effectively replicating the functionality of uneven pruning in HiNeRV~\cite{kwan2024hinerv}.

We incorporate this entropy estimation method in the training of the I-model and further extend it to the training of the P-model. As shown in Fig.~\ref{fig-RQAT}, for the $k$-th model ($k \% m > 0$), $\boldsymbol{\theta}_k^{\prime}$ and $\boldsymbol{\theta}_k^{*}$ refer to the parameters initialized from the previous model by the function $f_i$ and the optimized parameters via training process, respectively. We define the reference function as:
\begin{equation}
    f_c \left(\boldsymbol{\theta }_k^{\prime}, \boldsymbol{\theta }_k^{*} \right) = \boldsymbol{\theta }_k^{*} - \boldsymbol{\theta }_k^{\prime},
\end{equation}

While computing the reconstructed frame $\hat{\mathrm{v}}$, we use $\boldsymbol{\theta }_k^{*}$, but for entropy estimation of the network parameters, we use the difference $\boldsymbol{\theta }_k^{*} - \boldsymbol{\theta }_k^{\prime}$. The loss function is:
\begin{equation}
    \begin{cases}
	Loss_d=MSE\left( \mathrm{v},f\left( t;\boldsymbol{\theta }_{k}^{*} \right) \right),\\
	Loss_r=I\left( \boldsymbol{\theta }_{k}^{*}-\boldsymbol{\theta }_{k}^{\prime} \right),\\
	Loss=Loss_r+\lambda \cdot Loss_d,\\
    \end{cases}
\end{equation}
where $f$ denotes the forward process of an INR model parameterized by $\boldsymbol{\theta }_{k}^{*}$. This approach allows information from the previous model instance to improve convergence speed while simultaneously reducing redundancy.

We design this module in the principle of local temporal similarity within video sequences. It utilizes the model parameters of the preceding GOP as a reference for the current one. To enhance RD performance, our RQEC module establish a more accurate rate estimation model through consistent entropy estimation.

% 图一：不同random比例和PSNR的图
\begin{figure}[t]
\centering
\includegraphics[width=0.5\textwidth]{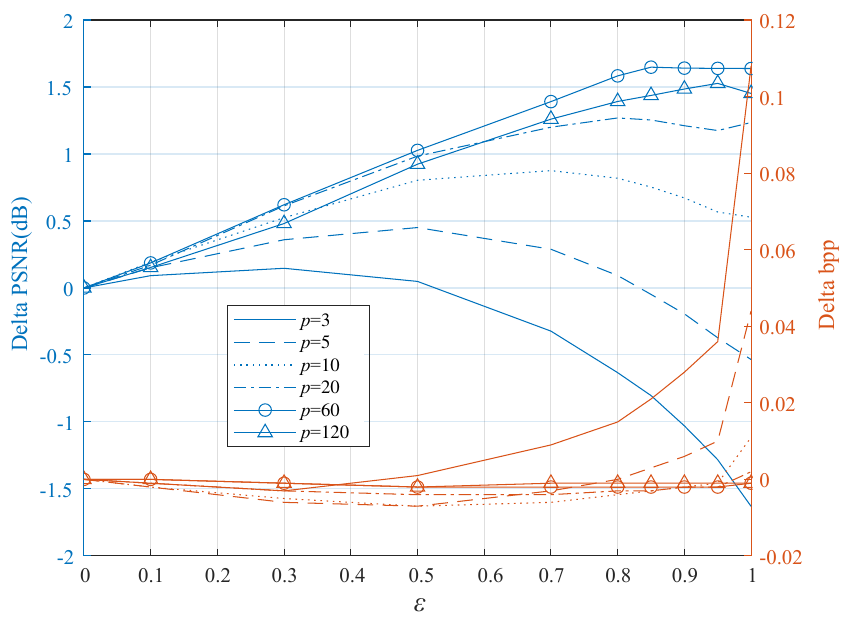}
\caption{Variation of RD performance with different random percent. 
The anchor for PNSR and bpp variations is $\varepsilon = 0$.
The legend for the right axis and the legend for the left axis share the same linetypes.
}
\label{fig-ReadySetGo-percent}
\end{figure}

% 图二：MCL-JCV上 percent-MSE*GOP的拟合图
\begin{figure}[t]
\centering
\includegraphics[width=0.5\textwidth]{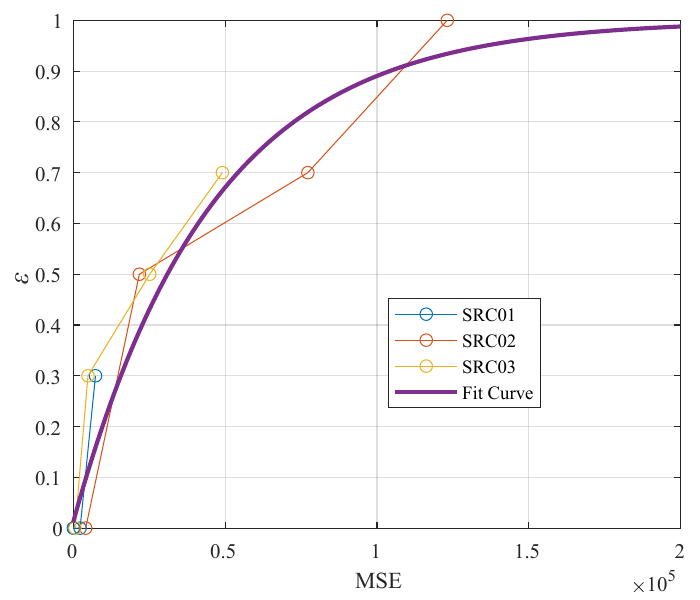}
\caption{Relationship between the gap of corresponding frames (measured using MSE) and the best random initialization percent $\varepsilon$.}
\label{fig-MCL-JCV}
\end{figure}

\subsection{Interpolation-based Model Initialization}
\label{sec-Interpolation-based Model Initialization}

Since adjacent frames in a video exhibit strong correlations, autoregressively initialized models that leverage parameters from neighboring model perform well when $p$ is small. However, as $p$ increases, these correlations weaken or even disappear. Consequently, the direct initialization (D-init) method may perform poorly, potentially even worse than random initialization (R-init). This phenomenon may arise from the large space of network parameters. The optimal parameters for GOP$\#a$ may be significantly different from those for GOP$\#b$, both of which are also distant from the randomly initialized parameters. In extreme cases, where the optimal parameters of two models lie in opposite directions, D-init can increase optimization difficulty rather than providing a useful starting point.

Based on this assumption, we propose an interpolation-based model initialization method, which dynamically controls the usage percentage of the trained parameters from the previous model. Specifically, we fuse preliminary parameters with a percentage factor $\varepsilon$ to regularize the trained parameters from the previous model. Our model initialization function is:
\begin{equation}
    \boldsymbol{\theta }_{k}^{\prime}=f_i\left( \boldsymbol{\theta }_{k-1},\boldsymbol{\theta }_{k-1}^{*} \right) =\varepsilon \cdot \boldsymbol{\theta }_{k-1}+\left( 1-\varepsilon \right) \cdot \boldsymbol{\theta }_{k-1}^{*},
\end{equation}
where $\varepsilon \in \left[ 0,1 \right]$, $ k \% m > 0$. This indicates that the interpolation initialization (II) module is not applied to the first model in a Group of Models (GOM). We only store a seed to reconstruct $\boldsymbol{\theta }_0$ on the decoding side, which serve as the initialization parameters of the first model in a GOM.

In Fig.~\ref{fig-ReadySetGo-percent}, we test the RD performance of HiNeRV on ``ReadySetGo" with different random percent, which indicates that the optimal $\varepsilon$ value corresponding to different gops is different.
To determine the optimal $\varepsilon$, we analyze the rate-distortion (RD) performance across different values of $\varepsilon$ and $p$. Our findings indicate that the best $\varepsilon$ varies depending on $p$. Specifically, as $p$ increases, the gap between adjacent GOPs widens, necessitating a larger $\varepsilon$ to achieve the best RD performance. This trend likely arises because a larger $p$ results in greater differences between corresponding frames in successive GOPs. To quantify the gap between two GOPs, we measure the Mean Squared Error (MSE) of corresponding frames~\cite{maiya2023nirvana}. To establish a numerical relationship between MSE and $\varepsilon$, we conduct a statistical analysis of the optimal $\varepsilon$ and MSE values across different $p$ settings using three sequences from the MCL-JCV~\cite{mcl-jcv-dataset} dataset, which are non-repetitive with the test datasets we use in Sec.~\ref{sec:experiments}. We assume this function follows a variant of an exponential function:
\begin{equation}
\label{eqn-percent}
    \varepsilon=-ae^{\left( -b \times MSE + c \right)}+1,
\end{equation}
where $a>0,b>0$. As illustrated in Fig.~\ref{fig-MCL-JCV}, we use the collected data to determine the hyperparameters $a,b$, and $c$, obtaining precise values. In all subsequent experiments on other datasets, we adopt the initialization function defined in Eqn.~(\ref{eqn-percent}).

The adaptive can be viewed as a simple, ``implicitly learned" meta-strategy, embodying the idea of ``learning how to initialize." It dynamically decides the extent to which past experience ($\boldsymbol{\theta }_{k-1}^{*}$) should be trusted and utilized, based on the similarity (measured by MSE) between the current task (encoding GOP k) and the previous task (encoding GOP k-1).

\section{EXPERIMENTS}
\label{sec:experiments}

In the following, we first present the experimental settings, followed by the overall rate-distortion performance of our framework. Next, we highlight the advantages of our framework under different $p$ settings from three perspectives. Finally, we conduct ablation studies to validate the effectiveness of our two modules.

\begin{figure*}[ht]
\centering
\includegraphics[width=1.0\textwidth]{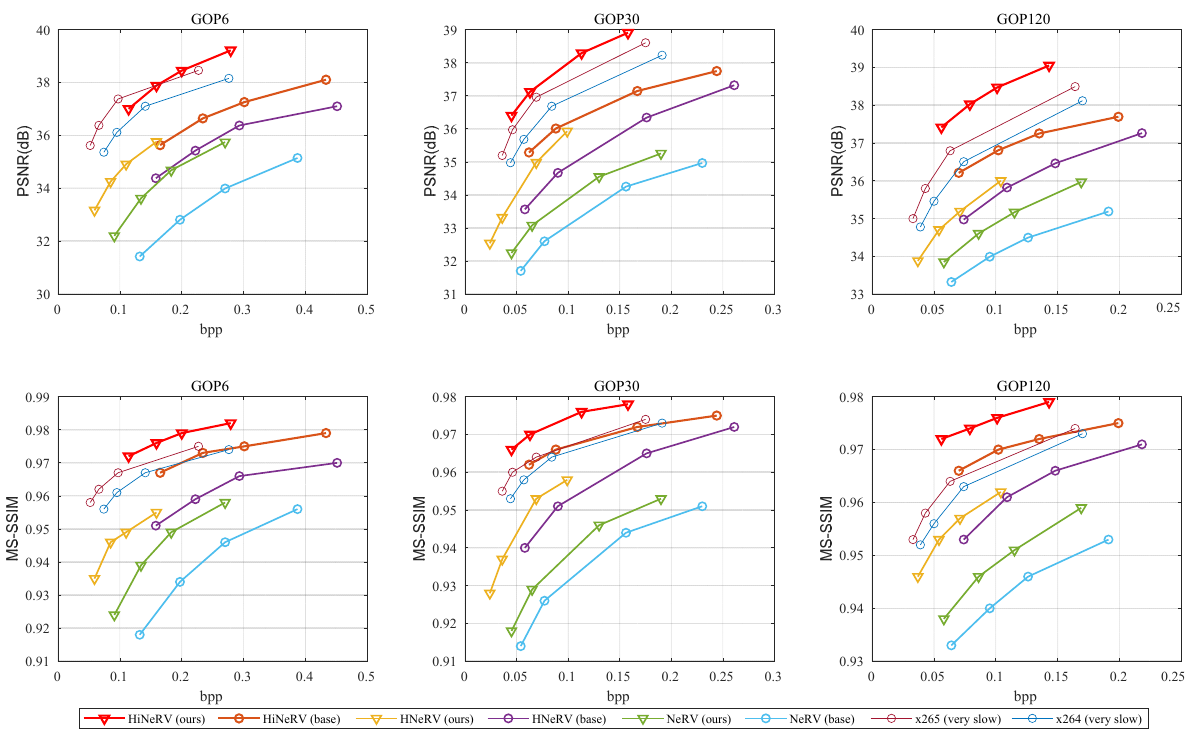}
\caption{Rate Distortion performance on UVG dataset with different $p$ (GOP).}
\label{fig-UVG-main_result}
\end{figure*}

% VVC ClassB result
\begin{figure*}[ht]
\centering
\includegraphics[width=1.0\textwidth]{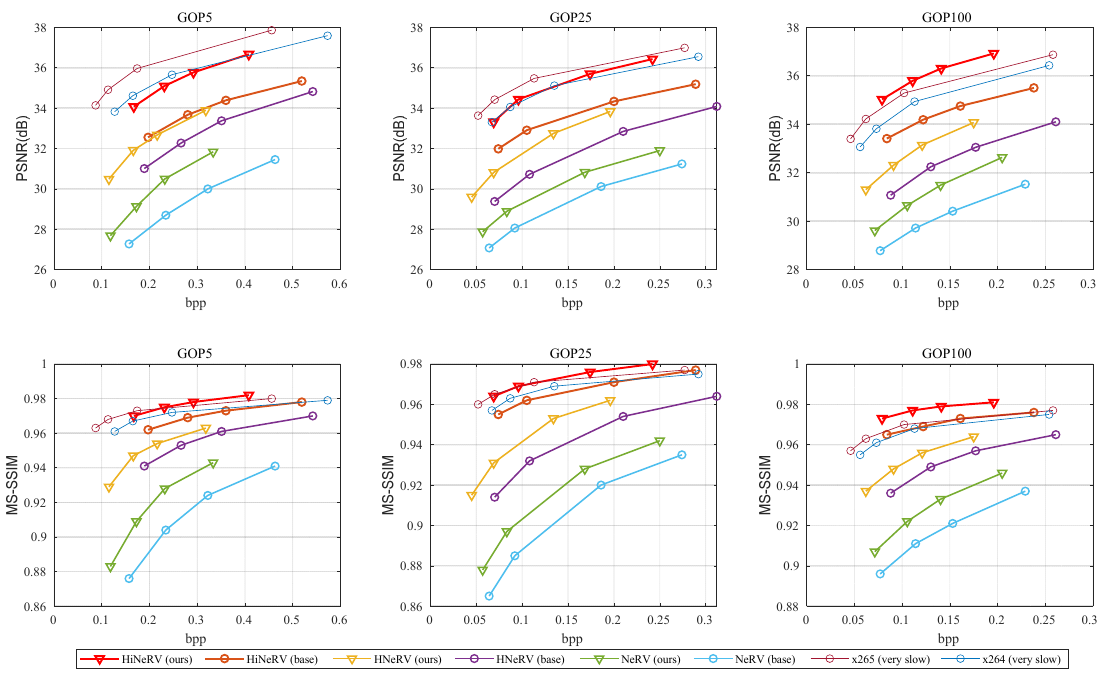}
\caption{Rate Distortion performance on VVC ClassB dataset with different $p$ (GOP).}
\label{fig-VVC-main_result}
\end{figure*}

\begin{table*}[]
\centering
\caption{BD-rate (\%) for PSNR/MS-SSIM on different datasets and $p$ (GOP).}
\label{bd-rate-table}
\resizebox{\textwidth}{!}{
\begin{tabular}{cc|cccccc|cccccc}
\hline
\multicolumn{1}{c|}{\multirow{3}{*}{Method}} & \multirow{3}{*}{Anchor} & \multicolumn{6}{c|}{UVG}                                                                                                                                                                                           & \multicolumn{6}{c}{VVC Class B}                                                                                                                                                                                    \\ \cline{3-14} 
\multicolumn{1}{c|}{}                        &                         & \multicolumn{3}{c|}{PSNR}                                                                                          & \multicolumn{3}{c|}{MS-SSIM}                                                                  & \multicolumn{3}{c|}{PSNR}                                                                                          & \multicolumn{3}{c}{MS-SSIM}                                                                   \\ \cline{3-14} 
\multicolumn{1}{c|}{}                        &                         & \multicolumn{1}{c|}{GOP6}            & \multicolumn{1}{c|}{GOP30}           & \multicolumn{1}{c|}{GOP120}          & \multicolumn{1}{c|}{GOP6}            & \multicolumn{1}{c|}{GOP30}           & GOP120          & \multicolumn{1}{c|}{GOP5}            & \multicolumn{1}{c|}{GOP25}           & \multicolumn{1}{c|}{GOP100}          & \multicolumn{1}{c|}{GOP5}            & \multicolumn{1}{c|}{GOP25}           & GOP100          \\ \hline
\multicolumn{1}{c|}{NeRV (ours)}             & NeRV (base)             & \multicolumn{1}{c|}{-45.04}          & \multicolumn{1}{c|}{-29.64}          & \multicolumn{1}{c|}{-35.99}          & \multicolumn{1}{c|}{-39.73}          & \multicolumn{1}{c|}{-24.15}          & -30.59          & \multicolumn{1}{c|}{-34.73}          & \multicolumn{1}{c|}{-30.99}          & \multicolumn{1}{c|}{-36.84}          & \multicolumn{1}{c|}{-31.92}          & \multicolumn{1}{c|}{-27.02}          & -32.29          \\ \hline
\multicolumn{1}{c|}{HNeRV (ours)}            & HNeRV (base)            & \multicolumn{1}{c|}{-41.08}          & \multicolumn{1}{c|}{-31.89}          & \multicolumn{1}{c|}{-11.79}          & \multicolumn{1}{c|}{-19.42}          & \multicolumn{1}{c|}{-30.26}          & -19.03          & \multicolumn{1}{c|}{-27.85}          & \multicolumn{1}{c|}{-36.18}          & \multicolumn{1}{c|}{-32.90}          & \multicolumn{1}{c|}{-22.78}          & \multicolumn{1}{c|}{-34.44}          & -28.64          \\ \hline
\multicolumn{1}{c|}{HiNeRV (ours)}           & HiNeRV (base)           & \multicolumn{1}{c|}{-58.94}          & \multicolumn{1}{c|}{-61.57}          & \multicolumn{1}{c|}{-64.97}          & \multicolumn{1}{c|}{-51.27}          & \multicolumn{1}{c|}{-53.74}          & -56.12          & \multicolumn{1}{c|}{-49.81}          & \multicolumn{1}{c|}{-52.24}          & \multicolumn{1}{c|}{-58.12}          & \multicolumn{1}{c|}{-44.11}          & \multicolumn{1}{c|}{-42.20}          & -53.54          \\ \hline
\multicolumn{2}{c|}{\textbf{Avg.}}                                     & \multicolumn{1}{c|}{\textbf{-48.35}} & \multicolumn{1}{c|}{\textbf{-41.03}} & \multicolumn{1}{c|}{\textbf{-37.58}} & \multicolumn{1}{c|}{\textbf{-36.81}} & \multicolumn{1}{c|}{\textbf{-36.05}} & \textbf{-35.25} & \multicolumn{1}{c|}{\textbf{-37.46}} & \multicolumn{1}{c|}{\textbf{-39.80}} & \multicolumn{1}{c|}{\textbf{-42.62}} & \multicolumn{1}{c|}{\textbf{-32.94}} & \multicolumn{1}{c|}{\textbf{-34.55}} & \textbf{-38.16} \\ \hline
\multicolumn{1}{c|}{HiNeRV (ours)}           & X265 (veryslow)         & \multicolumn{1}{c|}{20.43}           & \multicolumn{1}{c|}{-17.60}          & \multicolumn{1}{c|}{-34.29}          & \multicolumn{1}{c|}{-30.86}          & \multicolumn{1}{c|}{-51.29}          & -53.79          & \multicolumn{1}{c|}{85.55}           & \multicolumn{1}{c|}{36.34}           & \multicolumn{1}{c|}{-18.79}          & \multicolumn{1}{c|}{3.01}            & \multicolumn{1}{c|}{-4.82}           & -50.57          \\ \hline
\end{tabular}
}
\end{table*}

\subsection{Experiment Setting}

\subsubsection{Test Sequences and Base Models}
To validate the effectiveness of our framework, we conducted comprehensive evaluations on multiple models, including NeRV~\cite{chen2021nerv}, HNeRV~\cite{chen2023hnerv} and HiNeRV~\cite{kwan2024hinerv}. These tests spanned diverse datasets characterized by varied video content, resolution, and length:
\begin{itemize}
    \item UVG dataset~\cite{mercat2020uvg}: Six sequences (``HoneyBee", ``Bosphorus", ``Beauty", ``YachtRide", ``Jockey", ``ReadySetGo"), each comprising 600 frames at 1920x1080 resolution.
    \item VVC Class B dataset: Five sequences (``BasketballDrive", ``BQTerrace", ``Cactus", ``MarketPlace", ``RitualDance"), each containing 500 frames at 1920x1024 resolution.
    \item Big Buck Bunny: A video sequence of 132 frames with a 1280x720 resolution, used in NeRV.
\end{itemize}
We set $m=5$ and evaluate $p$ with values of (6, 30, 120) for the UVG dataset and (5, 25, 100) for the VVC Class B dataset to evaluate small, medium, and large $p$ scenarios. For bunny dataset, we set $m=22$ and $p\in(6, 22)$.
We provide two versions of the base models. The ``base" version employs the QAT~\cite{kwan2024hinerv} and D init. The ``ours" version incorporates our RQEC and II modules. 

\subsubsection{Implementation Details}
To achieve different bitrate points, we follow previous work to adjust the model size~\cite{chen2021nerv, chen2023hnerv, kwan2024hinerv}. For small $p$, the model sizes are (0.3, 0.45, 0.6, 0.9) M. For medium $p$, the model sizes are (0.6, 0.9, 1.8, 2.7) M. For large $p$, the model sizes are (3.0, 4.5, 6.0, 9.0) M. In addition, we set the rate-distortion trade-off parameter $\lambda$ to 5.0 for NeRV and HiNeRV, and 0.5 for HNeRV, as they have very different network architectures. 
We obtained these values using a fitting method similar to $\varepsilon$ described in Sec.~\ref{sec-Interpolation-based Model Initialization}.  
The small, medium and large $p$ settings correspond to different learning rates and training epochs for NeRV and HNeRV: 5e-3/5e-3/2e-3 and 3k/1.5k/1.5k for I-model, and 2k/2k/1k for P-model. For HiNeRV, the learning rates are adjusted to 2e-3/2e-3/1e-3. The papramter selection follows the setting in their original paper~\cite{kwan2024hinerv, chen2023hnerv}. Besides, we train for more epochs when $p$ is smaller, as fewer data points per epoch require longer training.

During training, we set the batch size to 1 and use the Adam optimizer. We also apply cosine learning rate decay with a 10\% warm-up, following previous NeRV-based approaches~\cite{chen2021nerv, chen2023hnerv, zhang2024boosting}. Most experiments are conducted on a single NVIDIA GTX 3090 GPU, except for some HiNeRV experiments, which require more memory and are trained on an NVIDIA A6000 GPU.

For traditional codecs, we compare our framework's performance with dominant video coding standards, including H.264~\cite{H.264} and H.265~\cite{H.265}. We also use the same value of $p$ as GOP for these traditional codecs for fair testing, which is (5/6, 25/30, 100/120).
We use x264 and x265 in FFmpeg with the veryslow profile as follows:

\begin{enumerate}[i)]
    \item \textit{ffmpeg -pix\_fmt yuv420p -s:v W$\times$H -i input.yuv -vframes N\_e -c:v libx264 -preset veryslow -qp QP -g GOP output.mkv}
    \item \textit{ffmpeg -pix\_fmt yuv420p -s:v W$\times$H -i Video.yuv -vframes Ne -c:v libx265 -preset veryslow -x265-params “qp=QP :keyint=GOP” output.mkv}
\end{enumerate}

\subsubsection{Evaluation Metrics}
We use Peak Signal-to-Noise Ratio (PSNR) and Multi-Scale Structural Similarity (MS-SSIM) to evaluate the quality of the reconstructed frames compared to the original frames. Bits per pixel (bpp) measures the average number of bits required to encode each pixel. In addition, we use the Bjøntegaard Delta Rate (BD-Rate)~\cite{bjontegaard2001BDBR} to compare the compression performance of different schemes, where negative values indicate bitrate savings, and positive values indicate increased bitrate consumption.

% Bunny result
\begin{figure*}[ht]
\centering
\includegraphics[width=1.0\textwidth]{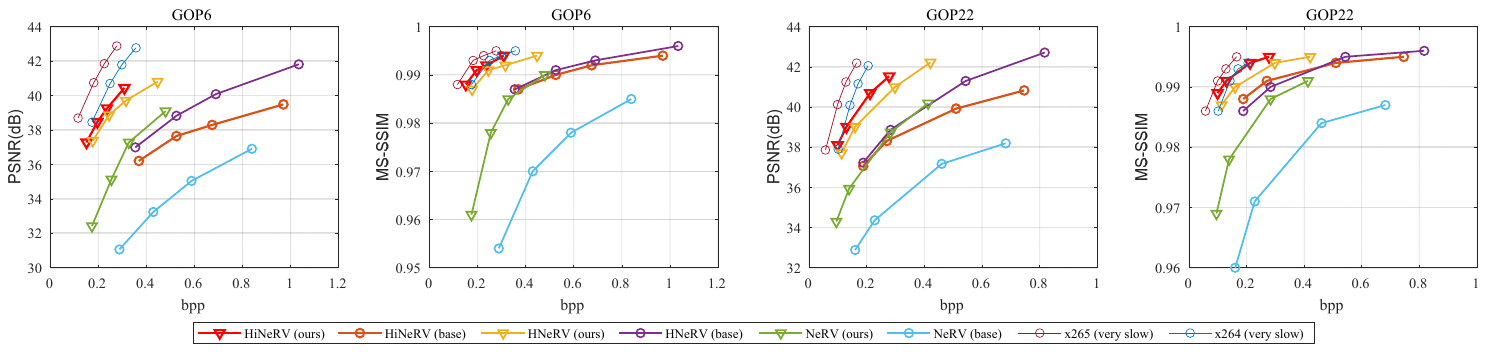}
\caption{Rate Distortion performance on Bunny dataset with different $p$ (GOP).}
\label{fig-Bunny-main_result}
\end{figure*}

\begin{table*}[]
\caption{Rate Distortion performance on different sequence of UVG.}
\label{table-sequence-RD}
\begin{tabular}{c|cccc|cccc|cccc}
\hline
GOP        & \multicolumn{4}{c|}{GOP6}                                                                                                                      & \multicolumn{4}{c|}{GOP30}                                                                                                                     & \multicolumn{4}{c}{GOP120}                                                                                                                     \\ \hline
Methods    & \multicolumn{2}{c|}{HiNeRV (base)}                         & \multicolumn{2}{c|}{HiNeRV (ours)}                                                & \multicolumn{2}{c|}{HiNeRV (base)}                         & \multicolumn{2}{c|}{HiNeRV (ours)}                                                & \multicolumn{2}{c|}{HiNeRV (base)}                         & \multicolumn{2}{c}{HiNeRV (ours)}                                                 \\ \hline
Metrics    & \multicolumn{1}{c|}{bpp}   & \multicolumn{1}{c|}{PSNR(dB)} & \multicolumn{1}{c|}{bpp}                          & PSNR(dB)                      & \multicolumn{1}{c|}{bpp}   & \multicolumn{1}{c|}{PSNR(dB)} & \multicolumn{1}{c|}{bpp}                          & PSNR(dB)                      & \multicolumn{1}{c|}{bpp}   & \multicolumn{1}{c|}{PSNR(dB)} & \multicolumn{1}{c|}{bpp}                          & PSNR(dB)                      \\ \hline
HoneyBee   & \multicolumn{1}{c|}{0.435} & \multicolumn{1}{c|}{38.697}   & \multicolumn{1}{c|}{0.104}                        & 39.136                        & \multicolumn{1}{c|}{0.246} & \multicolumn{1}{c|}{39.016}   & \multicolumn{1}{c|}{0.043}                        & 39.318                        & \multicolumn{1}{c|}{0.2}   & \multicolumn{1}{c|}{39.509}   & \multicolumn{1}{c|}{0.036}                        & 39.517                        \\ \hline
Bosphorus  & \multicolumn{1}{c|}{0.434} & \multicolumn{1}{c|}{41.083}   & \multicolumn{1}{c|}{0.231}                        & 42.229                        & \multicolumn{1}{c|}{0.245} & \multicolumn{1}{c|}{40.998}   & \multicolumn{1}{c|}{0.136}                        & 41.884                        & \multicolumn{1}{c|}{0.2}   & \multicolumn{1}{c|}{41}       & \multicolumn{1}{c|}{0.139}                        & 42.07                         \\ \hline
Beauty     & \multicolumn{1}{c|}{0.431} & \multicolumn{1}{c|}{35.265}   & \multicolumn{1}{c|}{0.362}                        & 35.94                         & \multicolumn{1}{c|}{0.242} & \multicolumn{1}{c|}{34.784}   & \multicolumn{1}{c|}{0.186}                        & 35.206                        & \multicolumn{1}{c|}{0.198} & \multicolumn{1}{c|}{34.657}   & \multicolumn{1}{c|}{0.177}                        & 35.199                        \\ \hline
YachtRide  & \multicolumn{1}{c|}{0.433} & \multicolumn{1}{c|}{37.468}   & \multicolumn{1}{c|}{0.348}                        & 39.443                        & \multicolumn{1}{c|}{0.244} & \multicolumn{1}{c|}{36.104}   & \multicolumn{1}{c|}{0.22}                         & 38.527                        & \multicolumn{1}{c|}{0.199} & \multicolumn{1}{c|}{35.587}   & \multicolumn{1}{c|}{0.208}                        & 38.856                        \\ \hline
Jockey     & \multicolumn{1}{c|}{0.432} & \multicolumn{1}{c|}{39.033}   & \multicolumn{1}{c|}{0.282}                        & 39.652                        & \multicolumn{1}{c|}{0.243} & \multicolumn{1}{c|}{38.788}   & \multicolumn{1}{c|}{0.158}                        & 39.343                        & \multicolumn{1}{c|}{0.198} & \multicolumn{1}{c|}{38.535}   & \multicolumn{1}{c|}{0.128}                        & 39.233                        \\ \hline
ReadySetGo & \multicolumn{1}{c|}{0.433} & \multicolumn{1}{c|}{37.096}   & \multicolumn{1}{c|}{0.346}                        & 38.915                        & \multicolumn{1}{c|}{0.244} & \multicolumn{1}{c|}{36.798}   & \multicolumn{1}{c|}{0.208}                        & 39.151                        & \multicolumn{1}{c|}{0.198} & \multicolumn{1}{c|}{36.885}   & \multicolumn{1}{c|}{0.17}                         & 39.441                        \\ \hline
Avg.       & \multicolumn{1}{c|}{0.433} & \multicolumn{1}{c|}{38.107}   & \multicolumn{1}{c|}{{\color[HTML]{FF0000} 0.279}} & {\color[HTML]{FF0000} 39.219} & \multicolumn{1}{c|}{0.244} & \multicolumn{1}{c|}{37.748}   & \multicolumn{1}{c|}{{\color[HTML]{FF0000} 0.158}} & {\color[HTML]{FF0000} 38.905} & \multicolumn{1}{c|}{0.199} & \multicolumn{1}{c|}{37.695}   & \multicolumn{1}{c|}{{\color[HTML]{FF0000} 0.143}} & {\color[HTML]{FF0000} 39.053} \\ \hline
\end{tabular}
\end{table*}

\subsection{Main Results on Rate-Distortion Performance}
\label{Rate-Distortion Performance}

\textbf{Overall performance with UVG dataset.}
To evaluate the performance of our framework under different settings, we first conduct experiments on the UVG dataset. As shown in Fig.~\ref{fig-UVG-main_result}, both NeRV, HNeRV and HiNeRV exhibit significant improvements when integrated with our proposed modules. Specifically, when using NeRV (base), HNeRV (base) and HiNeRV (base) as anchors, and comparing the RD curve positions of NeRV (ours), HNeRV (ours) and HiNeRV (ours), we observe that the primary gain in HNeRV stems from bitrate savings, while the primary gain in HiNeRV arises from improved image quality. At the same time, the structure of NeRV is more ordinary, improving a few in both dimensions.
This improvement is attributed to HiNeRV's more complex grid design, which enables it to capture finer details and fully leverage the potential of a more advanced entropy model with flexible bit allocation. Conversely, HNeRV's architecture is optimized for efficiency, featuring a compact structure that reduces memory consumption and accelerates inference speed. This streamlined design also imposes a limit on its maximum reconstruction quality, even as it achieves substantial bitrate reductions with our two modules.
When evaluated using the MS-SSIM metric, both models exhibit a trend similar to the analysis in PSNR. However, compared to x265 (veryslow) as the anchor, INR-based methods achieve significantly better in MS-SSIM relative to PSNR. This discrepancy arises because INR models encode video data in a more holistic manner, leading to superior structural fidelity. In contrast, traditional block-based encoders struggle to effectively capture the structural relationships between blocks.

\begin{figure*}[ht]
\centering
\includegraphics[width=1.0\textwidth]{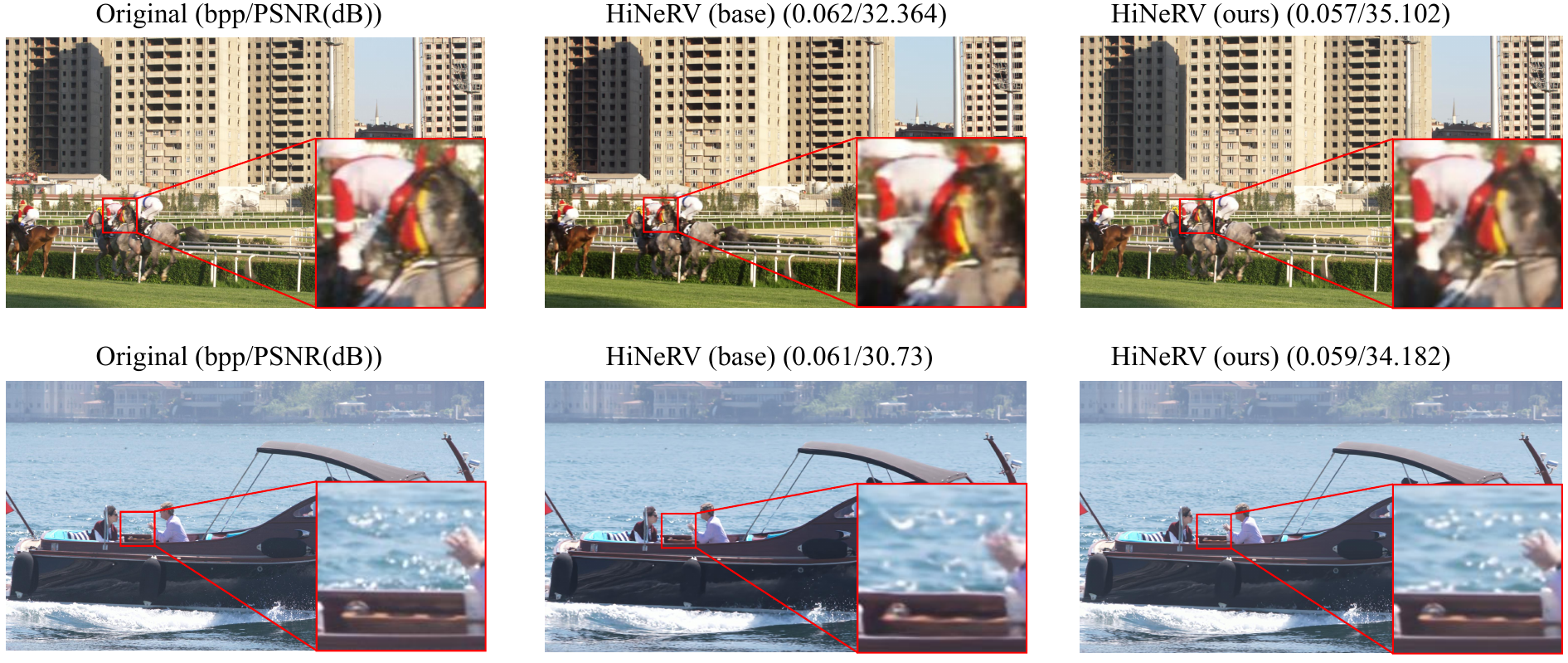}
\caption{Subjective quality comparison between HiNeRV (base) and HiNeRV (ours). HiNeRV (ours) achieves lower distortion at a similar bitrate.}
\label{selected_images}
\end{figure*}

To further illustrate the effectiveness of our method, Fig.~\ref{selected_images} presents visualized comparisons between HiNeRV (base) and HiNeRV (ours). The results clearly show that HiNeRV (ours) restores finer details while requiring a lower bitrate. Furthermore, HiNeRV (base) is more prone to amplifying certain details, causing them to become more prominent and interfering with the background. In contrast, HiNeRV (ours) models these details more effectively, preventing them from disrupting background consistency.

\textbf{Performance across difference $p$ settings.}
Examining the performance gap across different $p$ settings, we observe that in smaller $p$ settings, the RD performance of INR-based model is inferior to that in larger $p$ settings. This is primarily because models with larger $p$ settings are more effective at reducing temporal redundancy. Consequently, choosing an appropriate $p$ value involves a trade-off dictated by resource constraints.
In addition, we note that when $p=6$, HiNeRV (ours) performs less effectively than x265 (veryslow) at low bitrates but surpasses it at higher bitrates. This occurs because, when fitted with a smaller model, the HiNeRV structure has limited capacity to capture fine details, reducing its fitting ability. However, higher bitrate models alleviate this limitation, leading to improved overall performance.
As $p$ increases, the performance of our method improves steadily and eventually surpasses that of x265 (veryslow), demonstrating the robust modeling capabilities of implicit representation models. Notably, $p=30$ represents a highly practical configuration, with encoding delays that remain acceptable in many real-world scenarios. In this setup, our RD performance exceeds that of x265 (veryslow), further validating the effectiveness of our framework.
When $p=120$, HiNeRV (ours) significantly outperforms x265 (veryslow). This result demonstrates that our framework is not only suitable for lower-resource environments but is also capable of maximizing RD performance across a broad spectrum of resource configurations.

\textbf{Overall performance with VVC Class B dataset.}
Fig.~\ref{fig-VVC-main_result} presents the experimental results on the VVC Class B dataset. Most of the analyses observed on the UVG dataset, as discussed earlier, also apply here. However, due to the inherent characteristics of current INRs, which are more suitable for videos with fewer dynamic scenes, the overall gain relative to x265 (veryslow) on this dataset is lower than that observed on the UVG dataset. Nevertheless, at larger $p$ settings, the RD performance of HiNeRV (ours) still surpasses that of x265 (veryslow).

In addition, we present the BD-Rate results for both PSNR and MS-SSIM metrics in Table~\ref{bd-rate-table} to show the performance of HiNeRV (ours) over four bitrate points. Our approach has significant RD performance improvements across all base models. Among all INR-based methods, HiNeRV (ours) achieves the best performance. For the PSNR metric, it outperforms x265 (veryslow) in half of the experimental setups. In terms of MS-SSIM, HiNeRV (ours) exceeds x265 (veryslow) in most setups, achieving up to a 50\% bitrate reduction with maintaining equivalent MS-SSIM when $p=120$. 

In our framework, the dependency between different GOPs within the same GOM is relatively weak. Decoding a frame in the current GOP does not require the explicit prior decoding of frames from other dependent GOPs. As the resulting random-access latency is comparable to that of x264/x265 with a GOP size of p, we chose that setting for our primary comparison.
We also calculated the BD-Rate for PSNR against x265 (veryslow) with a GOP size of (p x m) on the UVG dataset. For HiNeRV (ours)-GOP6, the BD-rate relative to x265 changed from 20.43\% (vs. GOP 6) to 42.91\% (vs. GOP 6x5=30). For HiNeRV (ours)-GOP30, the BD-rate relative to x265 changed from -17.60\% (vs. GOP 30) to -15.51\% (vs. GOP 30x5=150). Although the performance margin is reduced under this comparison setting, the results for HiNeRV (ours) remain highly competitive, especially for longer GOPs.

\subsection{Performance on More Dimensions}
\textbf{Performance with difference video content.}
To validate the effectiveness of our framework on diverse video content, we present the RD performance of HiNeRV (base) and HiNeRV (ours) on various sequences from the UVG dataset. These include sequences with limited motion, such as ``HoneyBee", and those with significant motion, like ``ReadySetGo". According to the results in Table \ref{table-sequence-RD}, our framework demonstrates a significant performance advantage across all sequences. Notably, our RQEC module can dynamically adjust the optimization direction based on the RD behavior during training. For ``HoneyBee", a sequence where improving reconstruction quality is challenging, HiNeRV (ours) achieves substantial bitrate reduction. Conversely, for sequences with more room for quality improvement, such as ``ReadySetGo", HiNeRV (ours) significantly enhances reconstruction quality.

\textbf{Performance with difference video resolution (Bunny dataset).}
We also evaluated our framework's effectiveness on the Bunny dataset, with results presented in Fig.~\ref{fig-Bunny-main_result}. It is observed that HNeRV demonstrates better adaptability to such low-resolution video sequences, especially when $p$ is small. However, despite the inherent model architectures sometimes leading to performance characteristics that may not perfectly align with typical expectations across different configurations, our module consistently enhances the RD performance.

\textbf{Compare with other methods.} 
Having validated the flexibility of our framework and the effectiveness of its modules, we compare its performance against DCVC~\cite{li2021dcvc} and NIRVANA~\cite{maiya2023nirvana}. For DCVC, we calculated the BD-Rate of HiNeRV (ours) relative to DCVC under various $p$ settings, as presented in Table~\ref{table:DCVC}. Although our RD performance is comparatively lower at smaller $p$ values, it progressively surpasses DCVC as $p$ increases. Owing to our framework's flexibility, users can select different configurations based on their practical requirements and resource constraints. Regarding NIRVANA, we set $p$ to 3 and compare with the results it provided in the appendix on UVG (1920x1080), as no source code is provided. As shown in Table~\ref{table:NIRVANA}. While NIRVANA's original paper reported slightly superior RD performance over the original NeRV, our NeRV (ours) manages to outperform NIRVANA, even at this $p=3$ setting. Furthermore, HiNeRV (ours) demonstrates substantially superior RD performance. Despite NIRVANA employing a lightweight network architecture, our framework supports various NeRV models, allowing users to choose models optimized for speed or RD performance, without being restricted by a specific model structure.

\begin{table}[]
\centering
\label{table:DCVC}
\caption{The BD-rate (\%) of HiNeRV (ours) over DCVC, measured in PSNR.}
\begin{tabular}{c|c|c|c}
\hline
       & GOP6/5 & GOP30/25 & GOP120/100 \\ \hline
UVG    & 22.233 & -15.731  & -46.154   \\ \hline
ClassB & 80.427 & 25.169   & -20.093    \\ \hline
\end{tabular}
\end{table}

\begin{table}[]
\centering
\label{table:NIRVANA}
\caption{Detailed Comparison with NIRVANA over UVG.}
\begin{tabular}{c|cc|cc|cc}
\hline
           & \multicolumn{2}{c|}{NIRVANA}       & \multicolumn{2}{c|}{\begin{tabular}[c]{@{}c@{}}NeRV (ours)\\ GOP3-1.2M\end{tabular}} & \multicolumn{2}{c}{\begin{tabular}[c]{@{}c@{}}HiNeRV (ours)\\ GOP3-1.2M\end{tabular}} \\ \hline
           & \multicolumn{1}{c|}{PSNR}   & bpp  & \multicolumn{1}{c|}{RSNR}                       & bpp                                & \multicolumn{1}{c|}{RSNR}                            & bpp                            \\ \hline
HoneyBee   & \multicolumn{1}{c|}{38.83}  & 0.51 & \multicolumn{1}{c|}{39.624}                     & 0.52                               & \multicolumn{1}{c|}{\textbf{39.907}}                 & \textbf{0.366}                 \\ \hline
Bosphorus  & \multicolumn{1}{c|}{40.53}  & 0.68 & \multicolumn{1}{c|}{41.24}                      & 0.614                              & \multicolumn{1}{c|}{\textbf{43.353}}                 & \textbf{0.496}                 \\ \hline
Beauty     & \multicolumn{1}{c|}{35.77}  & 0.96 & \multicolumn{1}{c|}{35.576}                     & \textbf{0.646}                     & \multicolumn{1}{c|}{\textbf{37.577}}                 & 0.974                          \\ \hline
YachtRide  & \multicolumn{1}{c|}{37.94}  & 1.03 & \multicolumn{1}{c|}{37.339}                     & \textbf{0.761}                     & \multicolumn{1}{c|}{\textbf{41.741}}                 & 0.762                          \\ \hline
Jockey     & \multicolumn{1}{c|}{37.56}  & 0.96 & \multicolumn{1}{c|}{39.103}                     & \textbf{0.636}                     & \multicolumn{1}{c|}{\textbf{41.027}}                 & 0.71                           \\ \hline
ReadySetGo & \multicolumn{1}{c|}{35.43}  & 1.26 & \multicolumn{1}{c|}{35.234}                     & 0.782                              & \multicolumn{1}{c|}{\textbf{41.255}}                 & \textbf{0.745}                 \\ \hline
Avg.       & \multicolumn{1}{c|}{37.677} & 0.9  & \multicolumn{1}{c|}{38.019}                     & \textbf{0.66}                      & \multicolumn{1}{c|}{\textbf{40.81}}                  & 0.676                          \\ \hline
\end{tabular}
\end{table}

\subsection{Advantage of Different GOP Settings}
To achieve the same reconstruction quality for a given video dataset and model architecture, a smaller $p$ generally corresponds to a smaller model size, as fewer frames need to be fitted. Next, we analyze the impact of variations in model size.

\textbf{Memory Usage and Computational Complexity.}
Table~\ref{fig-memory-speed} presents the memory consumption of HiNeRV under different model sizes. We observe that as the model size increase, memory usage also rises. While the model sizes is lager than 9.0M, the training of HiNeRV could not be employed in an RTX 3090 GPU anymore.  More generally, our framework captures temporal correlations across multiple GOPs through autoregression, allowing each GOP to be loaded separately onto the GPU for training while maintaining an efficient compression ratio.

\textbf{Fitting Speed.}
Apart from low GPU memory consumption, another advantage of a small model is the reduction in forward-pass computation time, which impacts both training and inference speeds. Table~\ref{fig-memory-speed} presents the time required for training one frame (one forward and backward pass), showing that smaller model sizes result in faster processing speeds. 
Consequently, smaller $p$ settings ebables faster training. Early in training (not converged), the PSNR is higher for small $p$ with the same training time. In addition, the reduced computational cost during inference allows for operation on more lightweight devices, also making the approach suitable for Single-Encoding Multiple-Decoding (SEMD) scenario.

\begin{table*}[]
\centering
\caption{Detailed Memory Consumption and Speed Comparison across Base Models, Modules, and Model Sizes.}
\label{fig-memory-speed}
\begin{tabular}{c|c|c|c|ccc|cc|ccc}
\hline
\multirow{2}{*}{\begin{tabular}[c]{@{}c@{}}Model\\ Name\end{tabular}} & \multirow{2}{*}{\begin{tabular}[c]{@{}c@{}}Model\\ Size\end{tabular}} & \multirow{2}{*}{Module} & \multirow{2}{*}{Macs} & \multicolumn{3}{c|}{Encode}                                                                                                                                             & \multicolumn{2}{c|}{\begin{tabular}[c]{@{}c@{}}Entropy Coding of \\ Model Parameters\end{tabular}}                                           & \multicolumn{3}{c}{Decode}                                                                                                                                               \\ \cline{5-12} 
                                                                      &                                                                       &                         &                       & \multicolumn{1}{c|}{\begin{tabular}[c]{@{}c@{}}Memory\\ (MB)\end{tabular}} & \multicolumn{1}{c|}{\begin{tabular}[c]{@{}c@{}}Time per \\ Frame (s)\end{tabular}} & FPS   & \multicolumn{1}{c|}{\begin{tabular}[c]{@{}c@{}}Encode \\ Time (s)\end{tabular}} & \begin{tabular}[c]{@{}c@{}}Decode \\ Time (s)\end{tabular} & \multicolumn{1}{c|}{\begin{tabular}[c]{@{}c@{}}Memory\\ (MB)\end{tabular}} & \multicolumn{1}{c|}{\begin{tabular}[c]{@{}c@{}}Time per \\ Frame (s)\end{tabular}} & FPS    \\ \hline
NeRV                                                                  & 1.8M                                                                  & RQEC                    & 124.32G               & \multicolumn{1}{c|}{3508.67}                                               & \multicolumn{1}{c|}{0.07008}                                                       & 14.27 & \multicolumn{1}{c|}{0.1058}                                                     & 0.23411                                                    & \multicolumn{1}{c|}{1478.0}                                                & \multicolumn{1}{c|}{0.01336}                                                       & 74.85  \\ \hline
NeRV                                                                  & 1.8M                                                                  & QAT                     & 124.32G               & \multicolumn{1}{c|}{3464.0}                                                & \multicolumn{1}{c|}{0.0679}                                                        & 14.73 & \multicolumn{1}{c|}{0.46824}                                                    & 0.45308                                                    & \multicolumn{1}{c|}{1505.33}                                               & \multicolumn{1}{c|}{0.01316}                                                       & 75.99  \\ \hline
HNeRV                                                                 & 1.8M                                                                  & RQEC                    & 67.48G                & \multicolumn{1}{c|}{2966.0}                                                & \multicolumn{1}{c|}{0.06091}                                                       & 16.42 & \multicolumn{1}{c|}{0.11072}                                                    & 0.23413                                                    & \multicolumn{1}{c|}{1596.0}                                                & \multicolumn{1}{c|}{0.00766}                                                       & 130.55 \\ \hline
HNeRV                                                                 & 1.8M                                                                  & QAT                     & 67.48G                & \multicolumn{1}{c|}{2942.0}                                                & \multicolumn{1}{c|}{0.05932}                                                       & 16.86 & \multicolumn{1}{c|}{0.55829}                                                    & 0.44402                                                    & \multicolumn{1}{c|}{1412.0}                                                & \multicolumn{1}{c|}{0.00743}                                                       & 134.59 \\ \hline
HiNeRV                                                                & 1.8M                                                                  & RQEC                    & 125.55G               & \multicolumn{1}{c|}{11196.67}                                              & \multicolumn{1}{c|}{0.22136}                                                       & 4.52  & \multicolumn{1}{c|}{0.11698}                                                    & 0.27063                                                    & \multicolumn{1}{c|}{2923.33}                                               & \multicolumn{1}{c|}{0.08826}                                                       & 11.33  \\ \hline
HiNeRV                                                                & 1.8M                                                                  & QAT                     & 125.55G               & \multicolumn{1}{c|}{11020.0}                                               & \multicolumn{1}{c|}{0.21089}                                                       & 4.74  & \multicolumn{1}{c|}{0.94722}                                                    & 0.48933                                                    & \multicolumn{1}{c|}{2826.67}                                               & \multicolumn{1}{c|}{0.08755}                                                       & 11.42  \\ \hline \hline
HiNeRV                                                                & 0.3M                                                                  & RQEC                    & 21.19G                & \multicolumn{1}{c|}{5470.0}                                                & \multicolumn{1}{c|}{0.17695}                                                       & 5.65  & \multicolumn{1}{c|}{0.02995}                                                    & 0.08386                                                    & \multicolumn{1}{c|}{1415.33}                                               & \multicolumn{1}{c|}{0.07437}                                                       & 13.45  \\ \hline
HiNeRV                                                                & 0.3M                                                                  & QAT                     & 21.19G                & \multicolumn{1}{c|}{5370.67}                                               & \multicolumn{1}{c|}{0.16828}                                                       & 5.94  & \multicolumn{1}{c|}{0.6322}                                                     & 0.10451                                                    & \multicolumn{1}{c|}{1397.33}                                               & \multicolumn{1}{c|}{0.07371}                                                       & 13.57  \\ \hline
HiNeRV                                                                & 3.0M                                                                  & RQEC                    & 208.24G               & \multicolumn{1}{c|}{13838.0}                                               & \multicolumn{1}{c|}{0.29304}                                                       & 3.41  & \multicolumn{1}{c|}{0.19746}                                                    & 0.43358                                                    & \multicolumn{1}{c|}{5102.67}                                               & \multicolumn{1}{c|}{0.11662}                                                       & 8.57   \\ \hline
HiNeRV                                                                & 3.0M                                                                  & QAT                     & 208.24G               & \multicolumn{1}{c|}{13698.67}                                              & \multicolumn{1}{c|}{0.28437}                                                       & 3.52  & \multicolumn{1}{c|}{1.13121}                                                    & 0.85308                                                    & \multicolumn{1}{c|}{4480.0}                                                & \multicolumn{1}{c|}{0.11652}                                                       & 8.58   \\ \hline
HiNeRV                                                                & 6.0M                                                                  & RQEC                    & 413.21G               & \multicolumn{1}{c|}{19298.67}                                              & \multicolumn{1}{c|}{0.35207}                                                       & 2.84  & \multicolumn{1}{c|}{0.40545}                                                    & 0.84315                                                    & \multicolumn{1}{c|}{9161.33}                                               & \multicolumn{1}{c|}{0.13703}                                                       & 7.3    \\ \hline
HiNeRV                                                                & 6.0M                                                                  & QAT                     & 413.21G               & \multicolumn{1}{c|}{19244.67}                                              & \multicolumn{1}{c|}{0.34168}                                                       & 2.93  & \multicolumn{1}{c|}{1.84024}                                                    & 1.56098                                                    & \multicolumn{1}{c|}{8287.33}                                               & \multicolumn{1}{c|}{0.13649}                                                       & 7.33   \\ \hline
HiNeRV                                                                & 9.0M                                                                  & RQEC                    & 613.12G               & \multicolumn{1}{c|}{23596.67}                                              & \multicolumn{1}{c|}{0.50683}                                                       & 1.97  & \multicolumn{1}{c|}{0.66516}                                                    & 1.4059                                                     & \multicolumn{1}{c|}{12263.33}                                              & \multicolumn{1}{c|}{0.18056}                                                       & 5.54   \\ \hline
HiNeRV                                                                & 9.0M                                                                  & QAT                     & 613.12G               & \multicolumn{1}{c|}{23338.0}                                               & \multicolumn{1}{c|}{0.49542}                                                       & 2.02  & \multicolumn{1}{c|}{2.79206}                                                    & 2.37955                                                    & \multicolumn{1}{c|}{11045.33}                                              & \multicolumn{1}{c|}{0.18061}                                                       & 5.54   \\ \hline
\end{tabular}
\end{table*}

\subsection{Ablation Study}
\label{sec:ablation}

\begin{table}[]
\caption{Ablation study of UAR-NVC. Values in the table mean bpp/PSNR (dB).}
\label{table:Ablation table}
\begin{tabular}{l|c|c|c}
\hline
GOP                        & 6            & 30           & 120          \\ \hline
\begin{tabular}[c]{@{}c@{}}HiNeRV (ours) \\ (V0) II init + ref \end{tabular} & 0.279/39.219 & 0.158/38.905 & 0.143/39.053 \\ \hline
(V1) R init + w/o ref      & 0.327/38.758 & 0.174/38.503 & 0.145/38.927 \\ \hline
(V2) D init             & 0.401/37.786 & 0.226/37.289 & 0.185/37.785 \\ \hline
(V3) w/o ref               & 0.344/38.995 & 0.181/38.605 & 0.144/39.054 \\ \hline
\end{tabular}
\end{table}

\begin{figure}[t]
\centering
\includegraphics[width=0.50\textwidth]{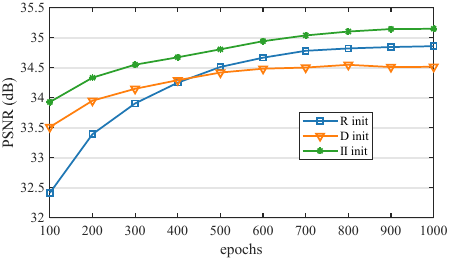}
\caption{The convergence process under the three initialization strategies.}
\label{fig-II_psnr}
\end{figure}

To evaluate the effectiveness of each proposed module, we conducted ablation experiments. We tested two components of UAR-NVC: the II module and the RQEC module. For the II module, we compared it against random initialization (R-init), which does not utilize any information from the trained parameters of previous models, and direct initialization (D-init), which directly transfers the trained parameters of the previous model as the initialization for the current model. For the RQEC module, we assessed performance in the absence of reference-based compression (w/o ref).

For $p=6/30$, we compare V0 and V3, as shown in Table~\ref{table:Ablation table}, and found that the reference compression design in the RQEC module reduces bitrate by 12.7\%/18.9\% while improving the PSNR metric by 0.3 dB/0.224 dB. This improvement occurs because the reference compression setting captures the correlation between adjacent GOPs, further enhancing reconstruction quality through the RD optimization mechanism of the RQEC module. Comparing V1 and V3, our II-init module provides only a marginal improvement. This is because, after approaching convergence, II-init and R-init have minimal impact on the final RD performance without reference.

Fig.~\ref{fig-II_psnr} illustrates the convergence process of the three initialization strategies. The results show that utilizing the similarity between GOPs can improve the reconstruction quality at the beginning of training.
As analyzed in Sec.~\ref{sec-Interpolation-based Model Initialization}, D-init leads to parameter overfitting to previous GOP data, trapping the optimization in local minima and restricting subsequent performance improvements.
In contrast, our II-init approach not only exploits the correlation between adjacent GOPs to accelerate training but also mitigates the overfitting issue observed in D-init.

For $p=120$, the performances of V0, V1, and V3 are nearly identical. This is because the performance gains derived from neighboring models primarily stem from the correlation between adjacent GOPs. However, at larger $p$ settings, this correlation diminishes, limiting the potential performance improvement.
Nevertheless, we observe that the two adverse effects caused by D-init persist, whereas our R-init effectively mitigates these issues.

To evaluate potential overhead introduced by our RQEC module, we split the whole process into encode (training), entropy coding and decode (inference), reporting training/inference speed, GPU memory usage, computational complexity, and entropy coding time under various settings in Table~\ref{fig-memory-speed}. We observe that, The RQEC module adds approximately 10\% to GPU memory usage during inference, which is an acceptable trade-off. For entropy coding, RQEC is markedly faster, which may be attributed to the use of different entropy coding libraries (the entropy coding library used by QAT in HiNeRV\footnote{https://pypi.org/project/torchac/} does not support the RQEC module well, so we used the constriction library\footnote{https://pypi.org/project/constriction/} instead). These findings are consistent across various base models and sizes, demonstrating that the RQEC module does not introduce substantial overhead.

\section{CONCLUSION AND FUTURE WORK}

In this work, we propose UAR-NVC, a neural video framework that unifies timeline-based and INR-based Neural Video Compression (NVC) from an autoregressive perspective, enabling current INR models to be adaptively utilized in different resource-constrained scenarios. Furthermore, to enhance rate-distortion (RD) performance, we introduce two key modules: The RQEC module, designed for network optimization and model compression; the II module, designed for efficient model initialization.

Although the framework has been introduced, several aspects require further improvement. Since the complexity of fitting videos varies significantly based on scene dynamics, future work should explore dynamic clip partitioning in the \textit{frame grouping} process to better adapt to different motion patterns.
In addition, our framework currently relies on existing INR models for clip processing, and the RD performance of UAR-NVC with small GOPs is not yet optimal. A more carefully designed and optimized INR model architecture could further enhance performance.
% Furthermore, our current approach to model reference compression and reference initialization is relatively simplistic. Future work could focus on refining these two modules to further improve efficiency and rate-distortion performance.
Finally, our current approach to model reference compression and reference initialization is relatively simplistic. Future work could focus on refining these two modules, like multi-GOP and multi-direction reference, to further improve efficiency and rate-distortion performance.

% \section*{Acknowledgments}
% This should be a simple paragraph before the References to thank those individuals and institutions who have supported your work on this article.

\bibliographystyle{IEEEtran}
\bibliography{reference.bib}

% \vfill

\end{document}